\documentclass[sigconf]{acmart}

\renewcommand\footnotetextcopyrightpermission[1]{} 

\AtBeginDocument{%
  \providecommand\BibTeX{{%
    \normalfont B\kern-0.5em{\scshape i\kern-0.25em b}\kern-0.8em\TeX}}}

\usepackage{multirow}
\usepackage{arydshln}
\usepackage{subfig}
\usepackage{tikz}
\usepackage{graphicx}

\newcommand{\wildcard}{\mathop{\scalebox{1.5}{\raisebox{-0.2ex}{$\ast$}}}}%

\begin{document}
	\fancyhead{}

\title[Segmenting Scientific Abstracts into Discourse Categories]{Segmenting Scientific Abstracts into Discourse Categories: A Deep Learning-Based Approach for Sparse Labeled Data}

\author{Soumya Banerjee}
\email{soumyaBanerjee@outlook.in}
\affiliation{     \institution{National Digital Library}   \institution{Indian Institute of Technology}   \city{Khragpur}   \state{India} }

\author{Debarshi Kumar Sanyal} 
\email{debarshisanyal@gmail.com}
\affiliation{     \institution{National Digital Library}   \institution{Indian Institute of Technology}   \city{Khragpur}   \state{India} }

\author{Samiran Chattopadhyay} 
\email{ samirancju@gmail.com}
\affiliation{     \institution{Department of IT}   \institution{Jadavpur University}   \city{Kolkata}   \state{India} }

\author{Plaban Kumar Bhowmick} 
\email{plaban@cet.iitkgp.ac.in} 
\affiliation{ \institution{Center for Education Technology}   \institution{Indian Institute of Technology}   \city{Khragpur}   \state{India} }

\author{Partha Pratim Das}
\email{ppd@cse.iitkgp.ac.in}
\affiliation{  \institution{Department of CSE}    \institution{Indian Institute of Technology}   \city{Khragpur}   \state{India}}

\renewcommand{\shortauthors}{Banerjee and Sanyal, et al.}

\begin{abstract}
The abstract of a scientific paper distills the contents of the paper into a short paragraph. In the biomedical literature, it is customary to structure an abstract into discourse categories like BACKGROUND, OBJECTIVE, METHOD, RESULT, and CONCLUSION, but this segmentation is uncommon in other fields like computer science. Explicit categories could be helpful for more granular, that is, discourse-level search and recommendation. The sparsity of labeled data makes it challenging to construct supervised machine learning solutions for automatic discourse-level segmentation of abstracts in non-bio domains. In this paper, we address this problem using transfer learning. In particular, we define three discourse categories -- BACKGROUND, TECHNIQUE, OBSERVATION -- for an abstract because these three categories are the most common. We train a deep neural network on structured abstracts from PubMed, then fine-tune it on a small hand-labeled corpus of computer science papers. We observe an accuracy of $75\%$ on the test corpus. We perform an ablation study to highlight the roles of the different parts of the model. Our method appears to be a promising solution to the automatic segmentation of abstracts, where the labeled data is sparse.
\end{abstract}


\maketitle

\section{Introduction}
The abstract of a research paper is a short, succinct description of the content of the paper. In a few lines, it conveys the information that is subsequently revealed in detail over multiple pages supplemented with figures,  tables, and references to existing works. In biomedical literature, including research papers, review articles, and clinical practice guidelines, it is a common practice to have structured abstracts \cite{imrad}. They typically follow the IMRaD format, i.e., INTRODUCTION, METHODS, RESULTS, and  DISCUSSION. 
Most medical journals indexed in PubMed conform to this style \cite{ripple2012structured}. Structured abstracts can help researchers to refer to their regions of interest quickly, label documents more effectively, assist the indexing process, and help in data mining \cite{ripple2012structured}. Recently researchers have designed deep network models to automatically segment unstructured abstracts in PubMed leveraging the large volume of structured abstracts already available \cite{dernoncourt2017pubmed, dernoncourt2017sequential, jin2018hierarchical}.  However, structured abstracts are uncommon in other disciplines like computer science, although, arguably, the same benefits may be reaped if structured abstracts were available (see, e.g., \cite{chan2018solvent}).

\definecolor{backgG}{RGB}{255, 255, 153}
\definecolor{tagtxtG}{RGB}{102, 102, 0}
\definecolor{backgPc}{RGB}{179, 255, 179}
\definecolor{tagtxtPc}{RGB}{0, 102, 0}
\definecolor{backgPw}{RGB}{255, 179, 179}
\definecolor{tagtxtPw}{RGB}{102, 0, 0}

\newcommand\goldentag[1]{%
	\tikz[baseline]{%
		\node[anchor=base, text=tagtxtG, fill=backgG, font=\sffamily, text depth=.005mm, rounded corners=0.08cm] {\tiny#1};
	}%
}
\newcommand\correcttag[1]{%
	\tikz[baseline]{%
		\node[anchor=base, text=tagtxtPc, fill=backgPc, font=\sffamily, text depth=.005mm, rounded corners=0.08cm] {\tiny#1};
	}%
}
\newcommand\wrongtag[1]{%
	\tikz[baseline]{%
		\node[anchor=base, text=tagtxtPw, fill=backgPw, font=\sffamily, text depth=.005mm, rounded corners=0.08cm] {\tiny#1};
	}%
}

\begin{figure*}[!htb]
	\centering 
	\small
	\begin{tabular}{ |p{17.5cm}|}
		\hline
		\textbf{Blending Digital and Face-to-Face Interaction Using a Co-Located Social Media App in Class } \\ \hline
		Improving face-to-face (f2f) interaction in large classrooms is a challenging task as student participation can be hard to initiate. \goldentag{BACKGROUND} \correcttag{BACKGROUND}\\
		Thanks to the wide adoption of personal mobile devices, it is possible to blend digital and face-to-face interaction and integrate co-located social media applications in the classroom. \goldentag{BACKGROUND} \correcttag{BACKGROUND}\\
		To better understand how such applications can interweave digital and f2f interaction, we performed a detailed analysis of real-world use cases of a particular co-located social media app: SpeakUp.\goldentag{TECHNIQUE} \correcttag{TECHNIQUE}\\
		In a nutshell, SpeakUp allows the creation of temporary location-bound chat rooms that are accessible by nearby users who can post and rate messages anonymously. \goldentag{TECHNIQUE} \wrongtag{OBSERVATION}\\
		We find that the use of co-located social media is associated with an increase in content-related interaction in the class.  \goldentag{OBSERVATION} \correcttag{OBSERVATION}\\
		Furthermore, it is associated with an increase in the perceived learning outcomes of students compared to a control group.  \goldentag{OBSERVATION} \correcttag{OBSERVATION}\\
		We further provide design guidelines to blend digital and f2f interaction using co-located social media in the classroom based on 11 case studies covering over 2,000 students. \goldentag{OBSERVATION} \correcttag{OBSERVATION}\\ \hline
	\end{tabular}
	\vspace{-6pt}
	\caption{An abstract, from \textit{cs.TLT} dataset,  with \colorbox{backgG}{golden} and predicted (\colorbox{backgPc}{correct} and \colorbox{backgPw}{errornous} ) labels.}
	\label{fig0}
\end{figure*}

In this paper, we investigate if sentences in an abstract can be labeled with discourse categories using machine learning methods even if the labeled data is sparse. We take abstracts in computer science (CS) as a case study. 
We categorize sentences in a CS abstract into three classes: BACKGROUND, TECHNIQUE, and OBSERVATION. 
We adopt a deep learning model for sequential sentence classification pretrained on structured abstracts from PubMed. We prepare a small corpus of hand-labeled abstracts in CS. We fine-tune the model on a subset of the corpus and test it on the remainder. Fig. \ref{fig0} shows an abstract from \textit{IEEE Transactions on Learning Technologies} in which each sentence is labeled with one of the three classes. Both hand-annotated golden labels and the predictions done by the model are indicated. 
We observe an accuracy of $75\%$ on our test corpus, which is quite promising, given the limited amount of golden data. In brief, our contributions are

\vspace{-.1cm}

\begin{enumerate}
	\item [1.] We propose a simplified discourse structure for CS abstracts.
	\item [2.] We prepare a hand-labeled corpus of structured CS abstracts.
	\item [3.] We use transfer learning to automatically classify sentences in CS abstracts into the above discourse categories.
\end{enumerate}
The code and the datasets are available publicly\footnote{https://github.com/soumyaxyz/abstractAnalysis}. 
The rest of the paper is structured as follows.  Section \ref{sec:discourseCategories} describes the discourse categories for CS abstracts and the rationale behind choosing this structure, Section \ref{sec:dataset} introduces the datasets, Section \ref{sec:model} describes the machine learning model, Section \ref{sec:eval}  evaluates of the model, and finally Section \ref{sec:conclusion} concludes the paper.

\section{Discourse Categories}
\label{sec:discourseCategories}

Abstracts in biomedical papers in PubMed are structured into five classes: BACKGROUND, OBJECTIVE, METHOD, RESULT, and CONCLUSION. As similar segmented CS abstracts are not available,  we employed human annotators to prepare our own datasets.
However, the annotators found it difficult to consistently annotate the sentences of CS abstracts into five distinct classes similar to those of the  PubMed abstracts.  Our discussions with the annotators revealed that the OBJECTIVE is hardly explicated; rather, the same sentence mixes the OBJECTIVE and the METHOD. These findings motivated us to adopt a compressed discourse structure where such sentences are labeled as TECHNIQUE. If a sentence is clearly an  OBJECTIVE, it has been labeled as BACKGROUND. Similarly, CS abstracts typically report some qualitative or quantitative findings without always making a general comment. Such sentences have been labeled as OBSERVATION. There can be a rare fourth sentence label, CODE, but the sentences in this class can be easily detected with regular expressions in a preprocessing step. Therefore, we remove such lines  from the corpus, before feeding the abstracts to the machine learning model.
Thus, a CS abstract is segmented into three classes: BACKGROUND, TECHNIQUE, and  OBSERVATION. The mapping between the two structures is shown in Table \ref{table2}. 

\vspace{-.3cm}
\begin{table}[!htb]
	\footnotesize
	\begin{tabular}{|l|r|}
		\hline
		BACKGROUND &     \multirow{2}{*}{BACKGROUND} \\ \cline{1-1}
		OBJECTIVE   &                                \\ \hline
		METHOD     &                  TECHNIQUE \\ \hline
		RESULT    & \multirow{2}{*}{OBSERVATION} \\ \cline{1-1}
		CONCLUSION &                                \\ \hline
	\end{tabular}
	\caption{Mapping between the discourse categories in biomedical and computer science papers.} \label{table2}
\end{table}
\vspace{-.95cm}

\section{Dataset}
\label{sec:dataset}
We introduce five new datasets of segmented abstracts in this paper. First, we introduce the \textit{PubMed-non-RCT} corpus containing abstracts of articles from PubMed that do not report randomized control trial (RCT). Like its RCT-analog in \cite{dernoncourt2017pubmed}, it contains 20k   abstracts, structured into the five classes already described.

\begin{table}[!htb]
	\footnotesize
	\begin{tabular}{| l l l l l|}
		\hline
		Dataset                 & Classes & Train        & Validation  & Test         \\ \hline
		\textit{PubMed-non-RCT} & 5       & 15k (165681) & 2.5k (26992) & 2.5k (24054) \\
		\textit{cs.NI}          & 3       & 110 (1224)   & -           & 40 (460)     \\
		\textit{cs.TLT}         & 3       & 110 (928)    & -           & 40 (326)     \\
		\textit{cs.TPAMI}       & 3       & 110 (901)    & -           & 40 (326)     \\
		\textit{cs.combined}    & 3       & 330 (3053)   & -           & 120 (1112)   \\ \hline
	\end{tabular}
	\caption{Dataset summary. Columns 3--5 show the number of abstracts in each subset, with the total number of lines in bracket.} \label{table1}
\end{table}

We created four corpora of CS abstracts: (1) \textit{cs.NI}: 150 abstracts on Networking and Internet Architecture from arXiv, (2) \textit{cs.TLT}: 150 abstracts from the journal \textit{IEEE Transactions on Learning Technologies}, (3) \textit{cs.TPAMI}: 150 abstracts from the journal \textit{IEEE Transactions on Transactions on Pattern Analysis and Machine Intelligence}, and (4) \textit{cs.combined}: the aggregation of the above three CS corpora. 
Unlike \textit{PubMed-non-RCT}, these abstracts were not available in structured format.
So, we first segmented each CS abstract into sentences with the spaCy\footnote{\texttt{https://spacy.io}} library. Each sentence in \textit{cs.TLT} and \textit{cs.TPAMI} abstracts was then labeled with one of the three classes by two independent annotators who are CS engineering graduates and, therefore, familiar with the domains. The inter-annotator agreement was found to be very high with Cohen's Kappa \cite{kappa} of 0.87 and 0.91, respectively. In case of \textit{cs.NI}, the dataset was distributed among senior CS  undergraduates. Subsequently, the annotations were reviewed by a CS doctoral scholar. Disagreements were resolved though mutual discussion. 
In a few instances, the sentence segmentation generated with spaCy was erroneous. The annotators corrected them manually.
The requirement of domain experts precluded the generation of very large labeled datasets in CS. Table \ref{table1} summarizes the datasets and how each is divided into train, validation, and test subsets.

\begin{figure}[htb]
	\centering
	\includegraphics[width=0.89\linewidth]{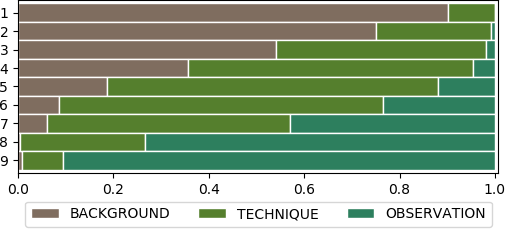}
	\vspace{-0.2cm}
	\caption{Distribution of labels in \textit{cs.combined}, visualized.} \label{fig4}
\end{figure}

\vspace{-.3cm}

The average number of lines in a CS abstract is approximately nine. Figure \ref{fig4} shows the distribution of labels in the \textit{cs.combined} corpus. Normalizing all abstracts to nine sentences, the figure demonstrates what fraction of which lines belong to which label.  It is apparent that the abstracts in the dataset roughly follow the label sequence: BACKGROUND $\rightarrow$ TECHNIQUE $\rightarrow$ OBSERVATION.

\section{Deep Learning Model for Discourse Category Identification}
\label{sec:model}
The proposed model utilizes the state-of-the-art sequential sentence classification architecture proposed by  Jin and Szolovits \cite{jin2018hierarchical}. 
Architecturally,  the model is composed of four conceptual layers.

\begin{itemize}
	\item[1)]The token processing layer: This layer, for a sentence of $n$  tokens $\{w_i, \cdots, w_n\}$, generates embeddings for each token and produces an $n\times D$ matrix, where $D$ is the embedding dimension. Pretrained GloVe word vectors \cite{glove} are used to initialize the input token vectors.
	
	\item[2)]The sentence processing layer: This layer accepts the output from the token processing layer and calculates the encoding for the whole sentence. The sentence encoding is calculated by passing the embedding matrix through a bidirectional LSTM and then applying self-attention on the output. 
	
	\item[3)]The abstract processing layer: Unlike the previous layers, this layer operates on a complete abstract at once. The layer accepts a matrix of $N$ vectors $\{s_i, \cdots, s_N\}$, where each $s_i$ is a sentence embedding. A bidirectional LSTM transforms the input matrix, which is then passed to a single dense layer. The output from this layer corresponds to the per-class score for each sentence.
	
	\item[4)]The output generation layer:  This final layer optimizes the prediction of the abstract processing layer by modeling the transition probability between two consecutive sentence labels with conditional random field (CRF) \cite{CRF}.
\end{itemize}

\begin{figure}[htb]
	\centering
	\includegraphics[width=0.99\linewidth]{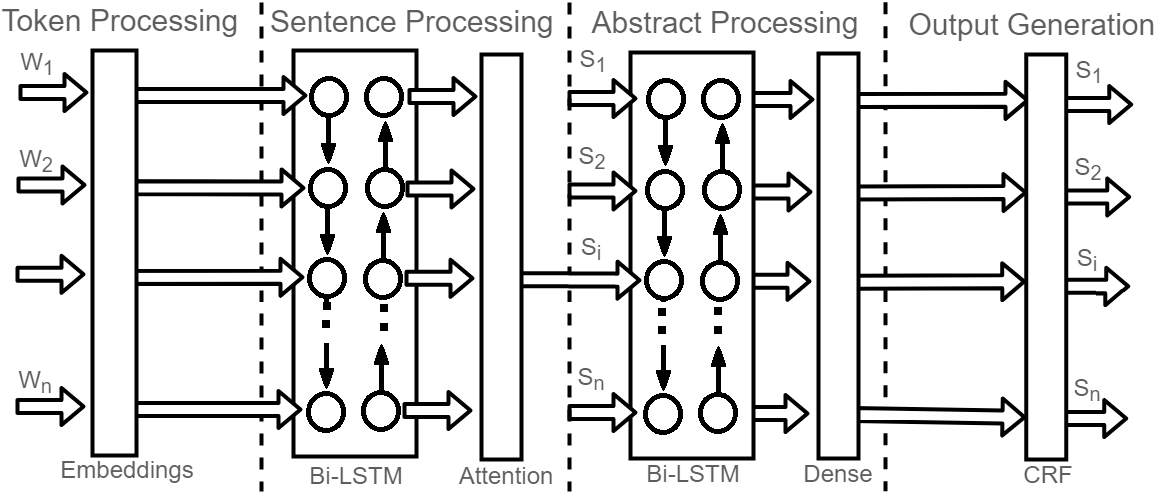}
	\caption{Model architecture.} \label{fig1}
\end{figure}
\vspace{-9pt}

The model architecture is summarized in Figure \ref{fig1}. During training, Adam optimizer  \cite{adam} and categorical cross-entropy loss function are applied. 
The model is first trained on the large  \textit{PubMed-non-RCT} corpus of structured biomedical abstracts. The resultant model is then fine-tuned on a much smaller CS dataset, i.e., the training subset of every \textit{cs.}$\wildcard$ dataset mentioned in Table \ref{table1}, and tested on the corresponding test subset.  

\section{Evaluation of the Model}
\label{sec:eval}
In this section, we evaluate the proposed model and validate our claims.

\subsection{Evaluation}
The \textit{PubMed-non-RCT} corpus was converted into a three-class dataset according to the mapping described in Table \ref{table2}. Then our proposed model was trained on it. Jin and Szolovits reported  $92.6\%$ accuracy on their \textit{PubMed 20k RCT} dataset. Our model achieves a comparable $92.1\%$  accuracy  on the biomedical \textit{PubMed-non-RCT} dataset. The model trained on \textit{PubMed-non-RCT} was saved. 
Next, we evaluated the model on each CS corpus listed in Table \ref{table1}.   
For each corpus, we evaluated in 3  ways, as shown in Table \ref{table3}: (1) \textit{Locally-trained:} The deep net model was trained only on the training subset and tested on the test subset of the CS corpus. 
(2) \textit{Pre-trained on PubMed:} The model was trained only on the \textit{PubMed-non-RCT} dataset and tested on the test subset of the CS corpus.  (3) \textit{Fine-tuned:} The model, pretrained on the \textit{PubMed-non-RCT} dataset, was fine-tuned on the training subset and tested on the test subset of the CS corpus. 
The results and the inferences are discussed in the following section.

\subsection{Results and Analysis}

Table \ref{table3} summarizes the accuracy of the locally trained model, the pre-trained model, and the fine-tuned model on all the four test subsets. Note that the accuracy of the model on a dataset is calculated as the percentage of sentences that are correctly labeled across all abstracts in the dataset.

\begin{table}[!htb]
	\scriptsize
	\begin{tabular}{| l l l|}
		\hline
		Dataset              & Details                              & Accuracy                         \\ \hline
		\textit{cs.NI}       & Locally trained                      & 54.08                            \\
		& Pre-trained on PubMed                & 29.46                            \\
		& Fine-tuned \phantom{---------------} & \textcolor{blue}{\textbf{65.22}} \\ \hline
		\textit{cs.TLT}      & Locally trained                      & 61.98                            \\
		& Pre-trained on PubMed                & 56.87                            \\
		& Fine-tuned                           & \textcolor{blue}{\textbf{79.45}} \\ \hline
		\textit{cs.TPAMI}    & Locally trained                      & 71.28                             \\
		& Pre-trained on PubMed                & 50.15                            \\
		& Fine-tuned                           & \textcolor{blue}{\textbf{83.44}} \\ \hline
		\textit{cs.combined} & Locally trained                      & 41.73                            \\
		& Pre-trained on PubMed                & 39.53                            \\
		& Fine-tuned                           & \textcolor{blue}{\textbf{75.18}} \\ \hline
	\end{tabular}
	\caption{Summary of results.} \label{table3}
\end{table}
\vspace{-16pt}

We can observe that transfer learning with fine-tuning ({`Fine-tuned'}) provides a significant improvement over the {`locally trained'} model. We also observe that without fine-tuning ({'Pre-trained on PubMed'}), transfer learning performs worse than local training. This indicates that the pretrained model is insufficient. It also gives an estimate of how dissimilar the particular CS corpus is compared to the biomedical corpus; if the CS corpus is similar to the biomedical corpus, further fine-tuning would be unnecessary.
The results indicate that \textit{cs.TLT} and \textit{cs.TPAMI} have modest similarity to \textit{PubMed-non-RCT} while \textit{cs.NI} has hardly any similarity as the $29.46 \%$ accuracy is as bad as a purely random guess.  Nevertheless, we  observe that, even when the datasets are completely dissimilar, transfer learning with fine-tuning on a marginal amount of labeled data can provide more than a $10\%$ accuracy boost.

\begin{figure}[htb]
	\centering
	\includegraphics[width=0.7\linewidth]{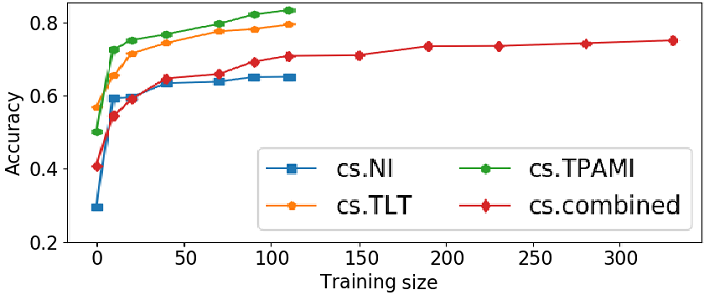}
	\caption{Effect of training size on accuracy.} \label{fig2}
\end{figure}

Figure \ref{fig2} demonstrates the effect of training size on the accuracy of the model on the four CS corpora. 
We observe that, initially the accuracy increases rapidly with the training size.
For comparatively larger size of the training corpus, accuracy still increases, albeit more gradually.

\captionsetup[subfigure]{labelformat=empty}
\begin{figure}[!ht]
	\centering
	\subfloat[][\textit{cs.NI}]{\includegraphics[width=.12\textwidth]{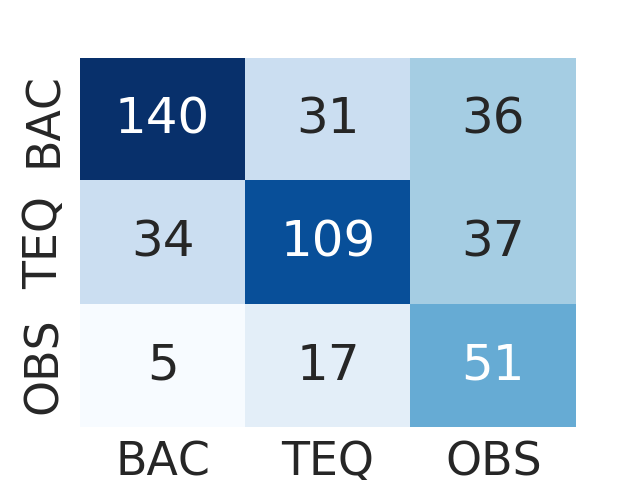}}
	\subfloat[][\textit{cs.TLT}]{\includegraphics[width=.12\textwidth]{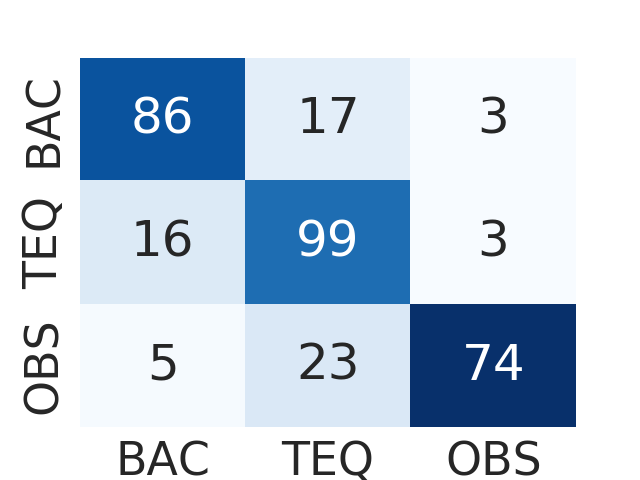}}
	\subfloat[][\textit{cs.TPAMI}]{\includegraphics[width=.12\textwidth]{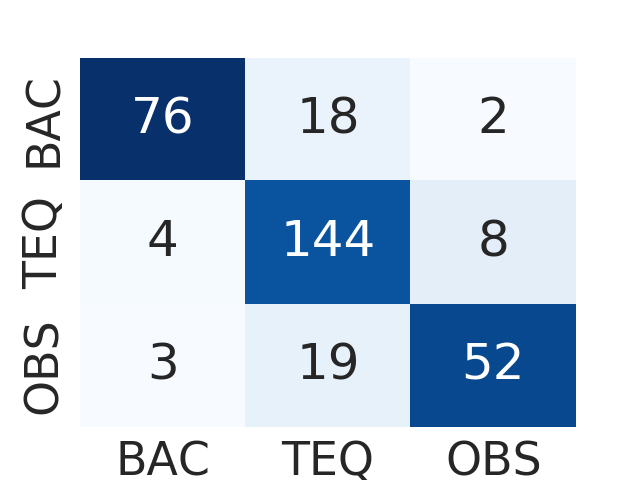}}
	\subfloat[][\textit{cs.combined}]{\includegraphics[width=.12\textwidth]{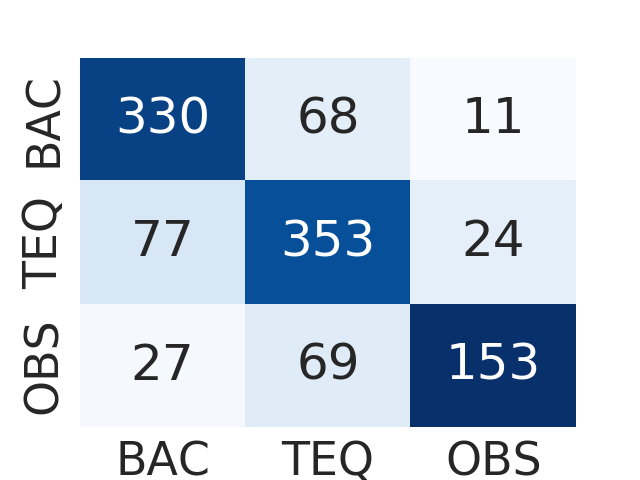}}
	\caption{Confusion matrices.}
	\label{fig3}
\end{figure}

Figure \ref{fig3} presents the confusion matrices  corresponding to the best runs (i.e., for the \textit{Fine-tuned} model) on the test subsets of the four CS corpora. Table \ref{table4} presents the per-class precision, recall, F1-score, and support, corresponding to the same runs.

\begin{table}[!htb]
	\footnotesize
	\begin{tabular}{|llllll|}
		\hline
		Dataset                               & Label       & P     & R     & F     & S   \\ \hline
		\multirow{3}{*}{\textit{cs.NI}}       & BACKGROUND  & 78.21 & 67.63 & 72.54 & 207 \\
		& TECHNIQUE   & 69.43 & 60.56 & 64.69 & 180 \\
		& OBSERVATION & 41.13 & 69.86 & 51.78 & 73  \\ \hline
		
		\multirow{3}{*}{\textit{cs.TLT}}      & BACKGROUND  & 80.37 & 81.13 & 80.75 & 106 \\
		& TECHNIQUE   & 71.22 & 83.90 & 77.04 & 118  \\
		& OBSERVATION & 92.50 & 72.55 & 81.32 & 102 \\ \hline
		
		\multirow{3}{*}{\textit{cs.TPAMI}}    & BACKGROUND  & 91.57 & 79.17 & 84.92 & 96  \\
		& TECHNIQUE   & 79.56 & 92.31 & 85.46 & 156 \\
		& OBSERVATION & 83.87 & 70.27 & 76.47 & 74  \\ \hline
		
		\multirow{3}{*}{\textit{cs.combined}} & BACKGROUND  & 76.04 & 80.68 & 78.29 & 409 \\
		& TECHNIQUE   & 72.04 & 77.75 & 74.79 & 454 \\
		& OBSERVATION & 81.38 & 61.45 & 70.02 & 249 \\ \hline
	\end{tabular}
	\caption{Results expressed in terms of (P)recision, (R)ecall, (F)1  and (S)upport.} 
	\label{table4}
\end{table}
\vspace{-14pt}


It is clear that the relatively poor accuracy for  \textit{cs.NI} is the result of the failure of our model to identify the sentences belonging to the class OBSERVATION.  In general, abstracts tend to follow the pattern BACKGROUND$\rightarrow$ TECHNIQUE$\rightarrow$  OBSERVATION. However,  as reported by the annotators, the abstracts in \textit{cs.NI} often deviate from this trend.
Moreover, \textit{cs.NI} is quite imbalanced, with significantly fewer instances of OBSERVATION. All these factors contribute to the model's difficulty with the \textit{cs.NI} dataset.   However, based on our observation in Fig \ref{fig2}, we expect with a larger volume of training data, these limitations can be overcome. 

\begin{table}[!htb]
	\footnotesize
	\begin{tabular}{| l c|}
		\hline
		Details                                                & Accuracy on \textit{cs.NI} test subset\\ \hline
		Fine-tuned on original \textit{cs.NI} & 65.22    \\
		Fine-tuned on noisy and augmented \textit{cs.NI}       
		&\textcolor{blue}{\textbf{67.99}}    \\ \hline
	\end{tabular}
	\caption{Experiment with noisy data.} \label{table5}
\end{table}
\vspace{-16pt}

To test this hypothesis, we performed an experiment. We took the model fine-tuned on \textit{cs.combined} dataset and used it to predict the labels for 150 additional abstracts from arXiv.CS.NI. These abstracts had no corresponding golden labels but we expect the predictions to have accuracy comparable to those of the test subset of \textit{cs.combined}. 
We augmented the  \textit{cs.NI} training subset, so that now it contains $110+150 = 260$ examples (albeit of somewhat questionable integrity). Then, the model that was pre-trained on PubMed was fine-tuned with this augmented corpus. We observed a $2\%$ increase in accuracy on the the original \textit{cs.NI} test subset (Table \ref{table5}). This vindicates our hypothesis.

\subsection{Ablation Analysis}
The ablation study of the proposed model in the context of \textit{cs.combined} is depicted in Table \ref{table6} where we subtract one layer at a time and report the changed accuracy. We observe that the abstract processing step, with the bidirectional LSTM to extract the conceptual information from the sentences, contributes the greatest toward the model's performance. Embeddings, with their capability to represent the contextual relationship between the tokens, has the second-highest contribution. The CRF in the output generation layer also has a major impact on the model's performance. This layer takes advantage of the sequential dependency in the output label sequence.  This ties into the poor performance on the \textit{cs.NI} dataset, as the sequential information is somewhat lacking in that dataset.

\begin{table}[!htb]
	\footnotesize
	\begin{tabular}{| l l |}
		\hline
		Dataset                                 & Accuracy \\ \hline
		\textit{Full Model}                     & 75.18       \\ \hdashline
		$-$ \textit{Token Processing}              & 63.56       \\
		$-$ \textit{Sentence Processing}           & 70.67       \\
		$-$ \textit{Abstract Processing}           & 43.15       \\
		$-$ \textit{CRF in Output Layer (replaced with softmax)}           & 66.02       \\ \hline
	\end{tabular}
	\caption{Summary of ablation analysis.} \label{table6}
\end{table}
\vspace{-18pt}

\section{Conclusion}
\label{sec:conclusion}
In this paper, we proposed a method for automatic discourse classification of sentences in computer science abstracts. We demonstrated that transfer learning with fine-tuning can provide remarkable results even on a sparsely labeled dataset.  We observed that due to the difference in presentation style, the nature of the discourse classification of CS abstracts vary across sub-fields. Nevertheless, the results on \textit{cs.combined} demonstrate that the proposed model generalizes fairly well across sub-fields of CS.

\begin{acks}
	This work is supported by the \textit{National Digital Library of India Project} sponsored by the Ministry of Human Resource Development, Government of India at IIT Kharagpur.
\end{acks}

\bibliographystyle{acm}
\bibliography{ref}


\end{document}